\documentclass[conference]{IEEEtran}
\usepackage{graphicx}
\usepackage{subcaption}
\usepackage{amsmath}
\usepackage{amssymb}
\usepackage{booktabs}
\usepackage{siunitx}

\title{Towards End to End Motion Planning and Execution for Autonomous Underwater Vehicles Using Reinforcement Learning}
\author{Elisei Shafer, Oren Gal \\
\small Department of Marine Technologies, University of Haifa, Israel}
\date{\today}

\begin{document}

\maketitle

\begin{abstract}
Autonomous Underwater Vehicles (AUVs) traditionally rely on complex, heavily engineered pipelines for perception, path planning, and motion control. This paper explores the feasibility of an end-to-end Deep Reinforcement Learning (DRL) approach that maps raw sensor data directly to thruster commands, reducing manual engineering. We propose a hierarchical reinforcement learning (HRL) architecture splitting the problem into two Markov Decision Processes. A High-Level (HL) policy operating at 2Hz processes raw $84 \times 84$ pixel monocular camera frames, stacked $100 \times 100$ pixel forward-looking imaging sonar, and proprioceptive data to generate spatial subgoals. Simultaneously, a Low-Level (LL) policy operating at 10Hz converts these subgoals into thruster commands. The HL policy is trained using Reinforcement Learning from Prior Demonstrations (RLPD) within a modified Sample-Efficient Robotic Reinforcement Learning (SERL) framework, while the LL policy utilizes Soft Actor-Critic (SAC) combined with Hindsight Experience Replay (HER). Evaluated in the high-fidelity HoloOcean simulator, our method demonstrates successful obstacle avoidance, achieving trajectory lengths closely approximating (within 4\% to 6\% of) an $\text{RRT}^*$ planning baseline. Furthermore, the learned policy exhibits strong robustness to simulated sensor noise and decreased visibility. While the system navigates familiar geometries effectively, experiments reveal generalization limitations when encountering unvisited areas with novel obstacle shapes. Ultimately, this work demonstrates the promise of sample-efficient, end-to-end DRL for underwater navigation using minimal computational hardware.
\end{abstract}

\section{Introduction}

AUVs have many uses in the marine environment: from research in the natural sciences to oil and gas and renewable energy. 
One of the more prevalent uses of AUVs is in area surveys, e.g. marine biology research to study an area, archeaolgical
surveys, surveys for construction and search and rescue.
One of the most basic subjects for AUVs are path and motion planning, obstacle avoidance and navigation. These are some of the most fundamental subjects in AUV research. Without it, an AUV cannot achieve any tasks, whether be it exploration, mapping, docking, etc.

Although these areas of research have been studied widely, most solutions require extensive engineering. For example, a survey of an unexplored area, close to the seafloor with obstacles would need a component for converting visual and sonar data into a map, a path planner/obstacle avoidance would then be applied to that map and then a controller would need to be implemented to follow said map. Each of these components needs to be implemented and configured. Our motivation in this paper is to explore the possibility of using minimal engineering/coding to create a policy that takes in raw sonar, visual and proprioceptive data and outputs commands to thrusters. We do this using deep reinforcement learning (DRL) methods and frameworks.

Deep reinforcement learning (RL) methods hold a promising avenue of research in solving the drawbacks of current methods, where an engineer would not need to engineer a solution, but rather the AUV would learn to avoid obstacles by itself or with the help of examples provided to it. We are inspired by the success of previous works such as champion level drone racing \cite{kaufmann2023drone_racing} and still other works that have trained robotic dogs \cite{daydreamer} and robotic arms for manipulation tasks \cite{serl} online, in the real world in record speed.

In this work we hope to show that end to end navigation can be achieved with DRL using raw sonar, proprioceptive and minimaly processed camera data as inputs and thruster output. We also hope to show that this can be done with a minimal amount of training of the policy - an amount such that the training can be done on a real vehicle in real time.

\section{Related Work}

\subsection{Traditional AUV methods}
The traditional AUV navigation pipeline involves multiple stages, from sensor data processing to thruster control. It encompasses perception, path planning, motion control, and integration challenges. AUVs utilize various sensors, including IMUs, depth sensors, cameras, and sonar, to navigate and inspect underwater structures \cite{Fernandes2015Pipeline} \cite{Sang2023Autonomous}. Navigation techniques include dead reckoning, signal-based navigation, and map-matching, often integrated for improved accuracy \cite{Zhang2023Autonomous}. Active perception frameworks can enhance navigation under sensor constraints \cite{Chang2022Active}. Real-time obstacle avoidance and visibility-aware motion planning are crucial for efficient operation \cite{Xanthidis2021AquaVis, Hernandez2016Autonomous}.

Underwater perception for autonomous navigation presents unique challenges due to limited visibility and GPS availability. Recent advancements in sensor fusion techniques have improved underwater SLAM by combining data from various sensors like IMUs, DVLs, cameras, and sonar \cite{Merveille2024Advancements}. Multi-sensor fusion approaches, including Kalman filters and particle filters, have been developed to improve navigation accuracy and robustness \cite{Nicosevici2004review}, \cite{Eustice2004Visually}.

\subsection{Deep Reinforcement Learning}
Early DRL approaches, such as those described by \cite{Noguchi2019Path}, combined artificial potential fields with reinforcement learning for seafloor tracking and intervention tasks. \cite{Xing2023Multi} integrated improved artificial potential fields with reinforcement learning (DDPG) and the traveling salesman problem to optimize energy efficiency and path cruising.
\cite{yuan2021auv} and \cite{Li2023Comprehensive} reported that reinforcement learning-based methods had higher success rates, path efficiency, and adaptability compared to genetic algorithms and traditional reinforcement learning in their simulation environments.

Visual sensors were less common but were integrated in studies such as \cite{Wu2019End}, where visual and sonar data were used for pipeline following and obstacle avoidance. \cite{Wu2019End} also highlighted the importance of integrating multiple sensor modalities to enhance robustness in noisy or uncertain environments.

There have been works that attempt end to end DRL in underwater robotics. \cite{Cai2024Learning} trained their policy on 2048 parallel simulations and use IMU, DVL and absolute location. \cite{Wu2019End} Created a two task system and used five pencil beam sonars for obstacle detection. \cite{Lyu2023End} also used pencil beam sonars in a two tier system, one tier handled planning and the other handled execution.

We note that in our review most implementations of DRL were on proprioceptive data (e.g. \cite{Cai2024Learning}). We found no works that use raw imaging sonar together with camera input. Works that included sonar were either pencil beam sonar \cite{Wu2019End, Lyu2023End, tang2024path} or divided imaging sonar data into bins \cite{tang2024path}, effectively turning the imaging into a pencil beam sonar array.

Traditional DRL algorithms such as PPO or even SAC take many agent/world interactions to train. Some works use highly parallel simulations to make training tractable but then have to deal with sim to real transfer \cite{Cai2024Learning}. Other works aim to use sample efficient algorithms such as HER \cite{Lyu2023End}. In other areas of robotics, even more advanced algorithms such as dreamer and RLPD\cite{rlpd} that have been shown to train a robot dog to walk \cite{daydreamer} and, using expert demos, to do advanced robotic manipulation tasks online on real hardware\cite{serl}.

\subsection{Our Contribution}
In our work we aim to show that it is possible to create an end to end policy using DRL. Although our experiments are in simulation, we also aim to show that our method can train a policy online within a feasible amount of both wall clock and simulation time, paving the way for future work to use our method online on a real AUV. While there are works that include DRL for AUV control, none that we know of use minimally processed camera and sonar data as input. Our method also uses minimal hardware: Training on a single Nvidia RTX 4000 16GB Ampere and inference working on an Nvidia RTX 3070 8GB gpu.

\section{End to End Deep Reinforcement Learning in an Underwater Setting}
\subsection{Problem Formulation}

We would like our AUV to go from starting pose $s$ and each a desired target goal $g$. 
The problem can be formulated as an MDP.
$\boldsymbol{s}, \boldsymbol{g} \subset \boldsymbol{\eta}$
\begin{equation}
    \boldsymbol{\eta}_s, \boldsymbol{\eta}_g \in \mathbb{R}^6
\end{equation}
Where the vehicle pose in fixed Earth coordinates are notated with $\boldsymbol{\eta}$

\begin{equation}
    \boldsymbol{\eta} = [x, y, z, \phi, \theta, \psi]
\end{equation}

Our work is done within the HoloOcean simulator which uses a right handed, forward left up (FLU) frame of reference for the AUV and fixed coordinates, hence we shall be using these coordinate systems in our formulation. Although we use FLU, we could easily swap the coordinate system with North East Down (NED) when training.

\subsection{Markov Decision Process}

A Markov Decision Process (MDP) provides the mathematical framework for modelling sequential decision-making problems in reinforcement learning. An MDP is defined by the tuple $(\mathcal{S}, \mathcal{A}, \mathcal{P}, \mathcal{R}, \gamma)$, where $\mathcal{S}$ is the state space, $\mathcal{A}$ is the action space, $\mathcal{P}$ is the transition probability function, $\mathcal{R}$ is the reward function, and $\gamma$ is the discount factor.

The MDP framework formalizes the agent-environment interaction: at each time step, the agent observes state $s_t$, selects action $a_t$, receives reward $r_t$, and transitions to state $s_{t+1}$. The agent's goal is to learn a policy $\pi(a|s)$ that maximizes the expected cumulative discounted reward $G_t = \sum_{k=0}^{\infty} \gamma^k r_{t+k}$.

\subsection{Algorithm Architecture Overview}
To make the problem tractable, we have decided to decouple motion execution from motion planning. We decided to implement this using hierarchial reinforcement learning (HRL), where we split the existing problem into two MDPs: One high level (HL) and the other low level (LL). The HL MDP has as its state camera, sonar and proprioceptive sensors, its action is the AUV's desired pose relative to its current pose which we define as a subgoal. It is relatively slow, with an update frequency of 2Hz, which allows it time to process raw 84x84 and 100x100 pixel camera and sonar images.
The LL MDP has only proprioceptive data and the subgoal as its state, its actions are the thruster commands. The LL policy works faster, at a rate of 10Hz, which allows it to be used as a motion controller.

\begin{figure}[h]
  \centering
  \includegraphics[width=\linewidth]{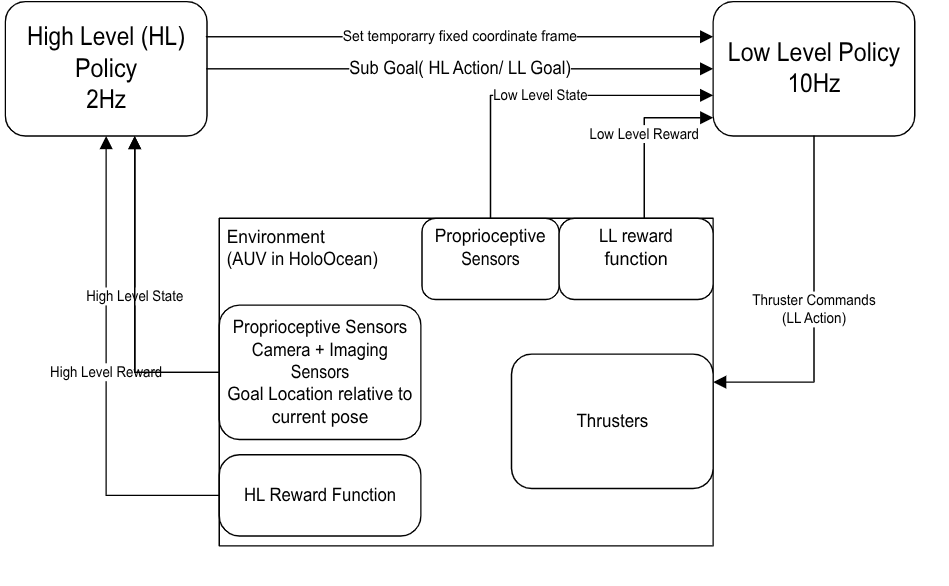}
  \caption[Hierarchial RL pipeline]{Hierarchial RL, diagram of the pipeline. We can see that the problem is split into two MDPs: High Level (HL) and Low Level (LL). The HL policy receives sonar, camera and prorioceptive data and its action is a subgoal, which is then set as the goal state of the LL policy. Both MPDs receive different outputs of the same environment as their state.}
  \label{fig:mdp_graph}
\end{figure}

\subsubsection{}{Low Level RL MDP}

\paragraph{State Space}
Our Proprioceptive State $\boldsymbol{s}_{\text{prop}} \in \mathbb{R}^{39}$ is made of the pose $\eta$ relative to starting position, velocity $\nu$, linear acceleration $\dot{\nu}_{\text{lin}}$:

\begin{align}
\boldsymbol{\eta} &= [x, y, z, \phi, \theta, \psi] \in \mathbb{R}^6 \\
\boldsymbol{\nu} &= [u, v, w, p, q, r] \in \mathbb{R}^6 \\
\dot{\boldsymbol{\nu}}_{\text{lin}} &= [\dot{u}, \dot{v}, \dot{w}] \in \mathbb{R}^3
\end{align}
We add goal states $g_{\text{desired}}$ and ${g}_{\text{achieved}}$ for goal conditioned learning and bound our desired goal in cube with edge length of 4m:
\begin{align}
\boldsymbol{g}_{\text{desired}} &= [\boldsymbol{\eta}_{\text{des}}, \boldsymbol{\nu}_{\text{des}}] \in [-2,2]^3 \times [-\pi, \pi] \times \mathbb{R}^{9} \\
\boldsymbol{g}_{\text{achieved}} &= [\boldsymbol{\eta}, \boldsymbol{\nu}] \in  \mathbb{R}^{12}
\end{align}
where $\boldsymbol{\nu}_{\text{des}} = [0,0, 0, 0, 0, 0]$ (twist is zero). Our combined proprioceptive state for the low level RL is the following:

\begin{equation}
\boldsymbol{s}_{\text{LL}} = [\boldsymbol{\eta}, \boldsymbol{\nu}, \dot{\boldsymbol{\nu}}_{\text{lin}}, \boldsymbol{g}_{\text{desired}}, \boldsymbol{g}_{\text{achieved}}]
\end{equation}

\paragraph{Action Space $\mathcal{A}$}

The action space consists of normalized thrust commands in four degrees of freedom:

\begin{equation}
\boldsymbol{a}_{\text{LL}} = [T_x, T_y, T_z, T_{\psi}] \in [-1, 1]^4
\end{equation}

where:
\begin{itemize}
    \item $T_x \in [-1, 1]$: Normalized surge (forward/backward)
    \item $T_y \in [-1, 1]$: Normalized sway (left/right)  
    \item $T_z \in [-1, 1]$: Normalized heave (up/down)
    \item $T_{\psi} \in [-1, 1]$: Normalized yaw (turn left/right)
\end{itemize}

The AUV has 8 thrusters, we convert the output from the policy to thruster values in the following way:

\begin{equation}
\left\{
\begin{aligned}
    T_{1,2,3,4} &= T_z \\
    T_5 &= T_x - T_{\psi} + T_y \\
    T_6 &= T_x - T_{\psi} - T_y \\
    T_7 &= T_x + T_{\psi} - T_y \\
    T_8 &= T_x + T_{\psi} + T_y
\end{aligned}
\right.
\end{equation}


\begin{figure}[h]
  \centering
  \includegraphics[width=\linewidth]{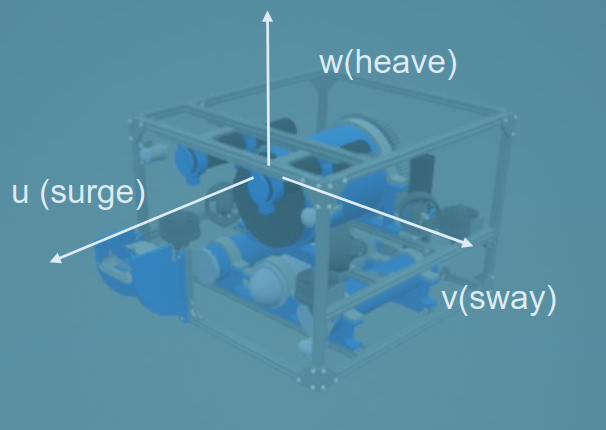}
  \caption{AUV coordinate frame.}
  \label{fig:coords}
\end{figure}

\subsubsection{High Level RL MDP }
The high level RL MDP is similar to the low level MDP with the following exceptions. The coordinate system is in the vehicle body frame of reference, thus we can drop the pose vector and the achieved goal. Since both would be zero. The desired goal is set to the desired pose of the AUV w.r.t. the current pose and is updated at each step relative to the AUV body frame. The action space is a 4 dimensional vector$(x,y,z,\psi)$ which is fed into the low level controller.

\paragraph{State Space}
The proprioceptive State for the high level controller, $\boldsymbol{s}_{\text{prop\_hl}} \in \mathbb{R}^{21}$ is the same as for the low level controller, but without ${g}_{\text{achieved}}$:

\begin{equation}
\boldsymbol{s}_{\text{prop}} = [\boldsymbol{\nu}, \dot{\boldsymbol{\nu}}_{\text{lin}}, \boldsymbol{g}_{\text{desired}}]
\end{equation}
The Exteroceptive State $\boldsymbol{s}_{\text{ext}}$ is made up of a Monocular Camera RGB image providing visual information about the underwater environment:
\begin{equation}
\boldsymbol{I}_{\text{cam}} \in [0, 255]^{84 \times 84 \times 3}
\end{equation}
We also stack the last 3 images from Forward-Looking Sonar(FLS):
\begin{equation}
\boldsymbol{I}_{\text{sonar}} \in \mathbb{R}^{100 \times 100 \times 3}
\end{equation}
The complete state space is of the following dimension:

\begin{align}
    \boldsymbol{s}_\text{HL} &= (\boldsymbol{s}_{\text{prop}}, \boldsymbol{I}_{\text{cam}}, \boldsymbol{I}_{\text{sonar}}) \\
    \boldsymbol{s}_\text{HL} &\in \mathbb{R}^{21} \times [0, 255]^{84 \times 84 \times 3} \times \mathbb{R}^{100 \times 100 \times 3}
\end{align}

\paragraph{Action Space}
The action space is a 4 dimensional vector$(x,y,z,\psi)$ which is the relative position and yaw orientation to the current position. We set positions to be a maximum of 1.9m.
\begin{align}
    \mathbf{a}_{\text{HL}} &= \boldsymbol{\delta} = [\Delta x, \Delta y, \Delta z, \Delta \psi] \\
    \text{s.t.} \quad & \boldsymbol{\delta} \in [-1.9, 1.9]^3 \times [-\pi, \pi]
\end{align}

\subsection{SERL}
To train the High Level (HL) reinforcement learning policy, this work utilizes a modified version of SERL \cite{serl}, a software suite specifically designed for sample-efficient robotic learning. Although SERL was originally developed for robot arm manipulation, its core architecture provides powerful tools for training policies in complex environments. The framework relies on the Reinforcement Learning from Prior Demonstrations (RLPD) algorithm, an off-policy method that trains the policy using a combination of prerecorded expert demonstrations and data collected online \cite{rlpd}. By performing multiple policy updates for each individual simulation step, RLPD significantly increases sample efficiency, making the training process more tractable for high-dimensional tasks.

The visual processing pipeline within this framework employs a pretrained ResNet-10 model as a preprocessing network to interpret environmental data. Additionally, DrQ (Data Regularized $Q$-learning) is used for image augmentation during training to improve the agent's ability to generalize across varying visual conditions. Our primary modifications to the SERL suite involved replacing the default robotic arm environment with custom Gymnasium environment files. These custom files enable the framework to interface directly with the AUV simulation, allowing the policy to ingest raw camera and sonar imagery and output desired subgoals for the vehicle.

\subsection{Soft Actor Critic (SAC) and Hindsight Experience Replay (HER)}
For low level RL training we use SAC \cite{sac} together with HER \cite{andrychowicz2017her}. SAC is a deep RL algorithm from the off policy family of algorithms which use a replay buffer to store transitions during training. HER is a method to modify the retrieval of transitions from the replay buffer. Some of the retrieved transition goals are modified such that the goal is the state that the agent reached. This helps the agent train faster, especially in cases with sparse rewards.

\subsection{Rewards}
For the low level RL, the reward is sparse with a -1 penalty for each timestep and +1 whenever the AUV reaches within 0.2m and 0.2rad of the goal coordinates.
For High level RL, the reward is dense reward: minus the distance to the goal and minus the angle to the desired angle. It receives +1 upon reaching within 1.9m of the desired goal. We also have a heavy penalty for collisions.

\section{Simulation Environment}
This section details the simulation architecture, the software interface, the specific sensor configurations, and the methodology used to acquire human expert data for the training pipeline

\subsection{HoloOcean Simulator}
The primary simulation platform used in this work is HoloOcean, a high-fidelity simulator built upon Unreal Engine. HoloOcean is specifically designed for underwater robotics research, providing realistic physics and the capability to simulate complex sensor modalities such as imaging sonar. The simulator uses a right handed, Forward-Left-Up (FLU) frame of reference for the AUV and the fixed Earth coordinates. In this paper we shall focus a scenario with geometric shapes (see figure \ref{fig:scenario})

\subsection{Gymnasium Environment}
To interface the reinforcement learning (RL) agents with the simulation environment, this work utilizes Gymnasium, an open-source Python library that provides a standardized API for reinforcement learning. Gymnasium serves as the essential abstraction layer that formalizes the AUV navigation problem as a Markov Decision Process (MDP).

In this research, a custom Gymnasium environment and wrapper files were developed to wrap the HoloOcean simulator, facilitating the communication between the simulation's physics engine and the RL algorithms. This implementation allows for a modular data flow where the environment provides complex high-dimensional observations, including $84\times84$ pixel monocular camera frames and $100\times100$ pixel forward-looking sonar images, to the agent. Simultaneously, Gymnasium handles the execution of actions within the simulator, such as translating policy outputs into normalized thruster commands. By leveraging this standardized framework, the environment was efficiently reconfigured to support the distinct requirements of the High Level (HL) motion planning policy and the Low Level (LL) motion execution policy, ensuring compatibility with established RL suites like SERL and Stable Baselines 3.

\subsection{Human Expert Demonstration Acquisition}
\begin{figure}[h]
  \centering
  \includegraphics[width=\linewidth]{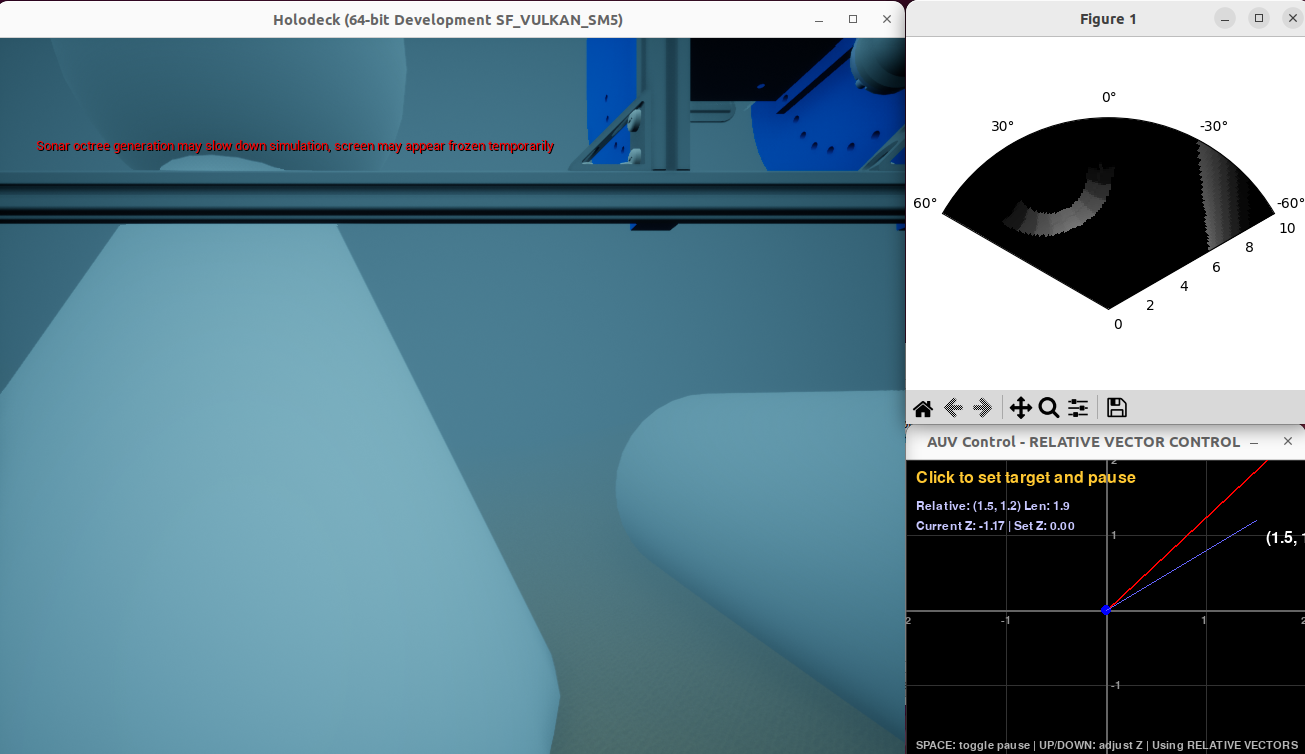}
  \caption[Human expert demonstration acquisition]{Human expert demonstration acquisition. The human receives an rgb image (center) and sonar camera data (top right). There is a screen with a line with the relative position of the goal (bottom right). The user can click on the screen which corresponds to the action for the high level RL.}
  \label{fig:demo}
\end{figure}

To facilitate training within the SERL framework, we developed a specialized tool to gather human expert demonstrations. During acquisition, a human operator observes the live camera and sonar feeds provided by the simulator (see figure \ref{fig:demo}). A dedicated interface displays the relative position of the goal as a visual cue. The operator provides inputs by clicking on relative positions in a control widget, which corresponds to the action space for the High Level (HL) RL policy. Additionally, the operator can control depth and yaw movements via a keypad. These recorded trajectories are saved as expert demonstrations to be used for Reinforcement Learning from Prior Demonstrations (RLPD).

\subsection{Sensor Simulation and constraints}
To ensure training stability and facilitate policy convergence within a feasible timeframe, several specific constraints were applied to the sensor simulations within the HoloOcean environment. First, all simulated sensors operate without noise, providing idealized data to the agent to simplify the initial learning process. Second, rather than implementing complex estimation algorithms, the AUV’s velocity and localization are provided by an "oracle" (perfect ground truth) directly from the simulator. The vehicle's proprioceptive suite provides linear and angular velocities ($\nu$), alongside linear accelerations ($\dot{\nu}_{lin}$), which are used for feedback in both high-level and low-level control loops. The exteroceptive suite consists of a monocular RGB camera and a forward-looking imaging sonar. For the high-level policy, the camera produces $84 \times 84$ pixel images, while the sonar input consists of the three most recent $100 \times 100$ pixel images stacked together to provide the temporal context necessary for the policy to interpret vehicle motion and obstacle proximity.

\section{Experiments}

\subsection{Baseline}
Finding a baseline for our algorithm was challenging since we found no previous works which combine raw sonar, camera and proprioceptive data in an end to end approach. We chose to compare our method with the following pipeline. A 2D obstacle map was given to the RRT* path planner, the RRT* path planner is then used to create a list of 2D waypoints without yaw orientation or depth. This list of waypoints is then fed one by one into a  proportional-derivative (PD) position controller.

\subsection{Training}
Training was done on consumer level hardware with an nVidia RTX4000 16GB graphics card.
\subsubsection{Low Level RL}
We ran training in an open water environment with no obstacles. The goal was set to a random location and yaw orientation in a 4x4x4m cube around the starting location of the AUV. Training was done using SAC+HER in Stable Baselines 3 \cite{stable-baselines3}. The policy converged after 150k steps or roughly 4 hours in simulation time. We then saved the best policy.
\subsubsection{High Level RL}
\begin{figure}[h]
  \centering
  \includegraphics[width=\linewidth]{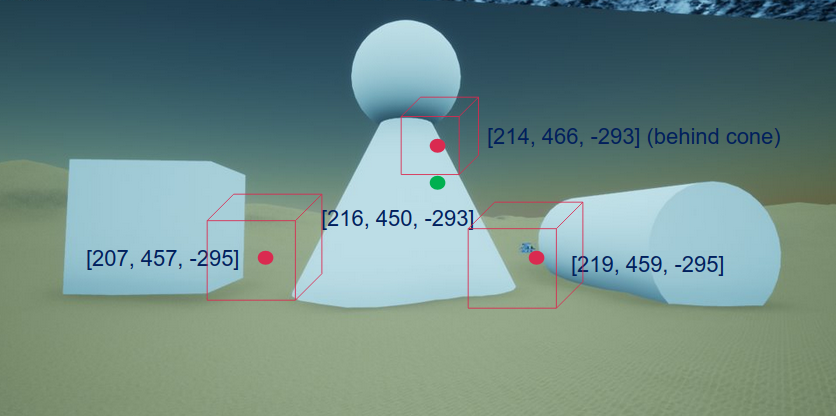}
  \caption[Training Scenario]{Training Scenario: The AUV starts at the green dot and is randomly assigned a goal in one of the areas(red cubes, not exact location) around the red dots. Note that fog has been removed for illustration purposes.}
  \label{fig:scenario}
\end{figure}

For our training scenario we chose a location in the HoloOcean Simulator OpenWater scenario with three large basic shapes: a square, cone, and cylinder. Our starting point was the same but we had 3 different areas for the goals (see figure \ref{fig:scenario}). The training goal areas were chosen such that the AUV would have to move inbetween obstacles and not just go above them, which in most situations is the easiest course of action.
We used SERL as our framework for training. Prior to starting, we recorded 80 demonstrations to all goal areas and in open water close to goals.

\subsection{Results}
\begin{figure}[htbp]
  \centering
  \begin{subfigure}[b]{0.48\textwidth}
    \centering
    \includegraphics[width=\linewidth]{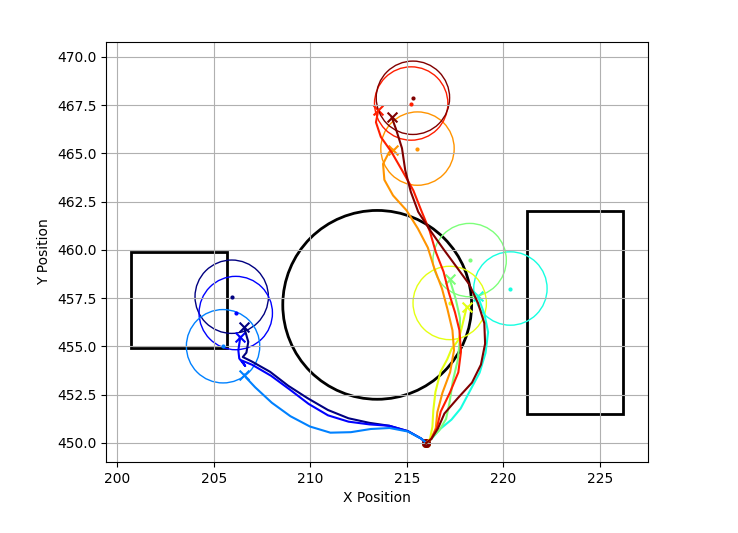}
    \caption{Test with goals in training (seen) areas: Top down view. Circles are desired goals, lines show the path of the AUV.}
    \label{fig:seen}
  \end{subfigure}
  \hfill
  \begin{subfigure}[b]{0.48\textwidth}
    \centering
    \includegraphics[width=\linewidth]{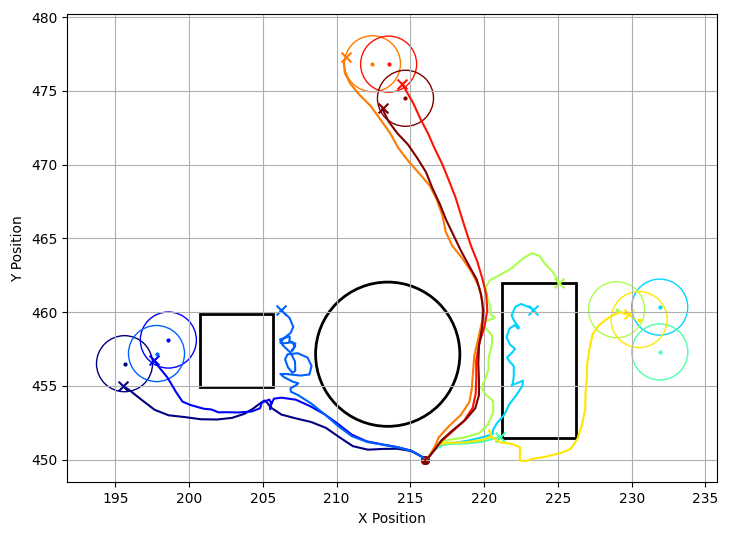}
    \caption{Test with previously unvisited (unseen) areas. The AUV enters a local minima behavior when encountering unseen states.}
    \label{fig:unseen}
  \end{subfigure}

  \caption{Comparison of trajectories in seen vs. unseen environments.}
  \label{fig:combined_auv_tests}
\end{figure}

For the low level policy, we ran 20 tests on random locations in the vicinity of the AUV. Success was measured if the AUV reached the within 20cm of the goal and within 11 degrees of the desired heading. We succeeded reaching the desired goal 20/20 times.
For the hl policy, we conducted 2 sets of tests: One set to the same areas as training (see figure \ref{fig:seen}), another to areas further away, to areas not seen in the training (see figure \ref{fig:unseen}). For the former, the policy performed well with the goals being reached 20/20 times for all test areas except behind the cone with 19/20. For the latter, the results were not as good:  Although the AUV could reach further away in empty space, 10m more than seen in tests with 20/20 successes, goals behind the cube had a success rate of 12/20 and the policy failed to generalize for the goal behind the cylinder with only 2/20 successes. We hypothesise that this is due to the shape of the cylinder, the end of which hasn't been seen in training and that during training there was no need to go above obstacles.
The significant performance degradation in the 'near cylinder' scenario (2/20 success rate)  suggests a limitation in the policy's structural generalization. While the agent successfully navigated obstacles with planar or conical surfaces seen during training, the convex hemispherical geometry of the cylinder’s end-cap represented an Out of Distribution (OOD) spatial feature. Since the training distribution lacked similar curvilinear primitives requiring vertical or lateral clearance, the policy failed to resolve the required maneuvers, leading to the observed local minima behavior. We have summarised the results in Table \ref{tab:tests}.

 Our experiments show that on visited areas, our policy is able to tell apart obstacles and avoid them. On unvisited areas, the policy is able to show good results and avoid obstacles so long as it has been trained on similar obstacles.

\begin{table}[htbp]
\centering
\caption[Methods and Coordinates Comparison]{Methods and Coordinates Comparison. Seen goals were from areas seen in the training set. Unseen goals were not seen in the training set.}
\label{tab:tests}
\begin{tabular}{|l|c|c|c|c|}
\hline
\textbf{Area type \textbackslash Area} & \textbf{Near Cube} & \textbf{Behind Cone} & \textbf{Near Cylinder} \\
\hline
Seen goals & 20/20 & 19/20 & 20/20 \\
\hline
Unseen goals & 12/20 & 20/20 & 2/20 \\
\hline
\end{tabular}
\end{table}

\subsubsection{RRT* Sampling Based Planner}
To compare the results we ran the following experiment using sampling based planning (SBP), specifically RRT*. The planner was written in Python and executed on a 2D map of the environment with rectangles and circles for obstacles. The planner is executed beforehand, the resulting path is then loaded to a path following script which feeds waypoints to a PD controller given in HoloOcean.

Since SBP and RRT* are not the main focus of this paper, we have chosen to use a simple low dimensional 2D planner with no planning for yaw. The hovering AUV is holonomic, which can follow a 2D path with a constant yaw. All paths were calculated with 5000 iterations.

We chose three goals that are located in the areas where the RL agent was trained: one between the cube and cone, one between the cone and cylinder and one behind the cone. For each goal, 3 paths were generated using RRT*, each path was executed by the follower 5 times for a total of 15 per goal. Also 10 runs for each exact goal were done using the RL agent. All trajectories were recorded at simulation update frequency, 10Hz, and statistics for each experiment were calculated. The results are consolidated in Table \ref{tab:trajectory_length}

Most RRT* results are as expected: the planner finds an efficient path, and the path follower follows the path. There had to be some adjustments made, notably the obstacles had to be inflated by 1m to account for robot size and for safety since the path follower does not follow the path perfectly.

All trajectory lengths from the RRT* with PD were shorter, except for the "behind cone" scenario where the actual obstacle was actually smaller than in the map for the specific depth. In the other areas where the map is very accurate, the trajectory length for our method is 1\% to 6\% longer than RRT* with PD. Where the lower percentage is for the areas seen during training on and the higher percentages for the "far behind cone" scenario (which was not seen in training)

This comparison is not without flaws. Firstly, the map given to the RRT* planner is relatively complete, i.e. the shape and location of obstacles are known beforehand, whereas the RL planner only has the information that it currently senses and some history. For RRT*, the map of the objects given to it are not the correct size for any given depth, e.g. when going behind the cone as in fig \ref{fig:seen} the RL agent goes through the obstacle on the map since it represents the base of the obstacle and not the actual size whereas the RRT* planner will plan around the base.

\begin{figure}[h]
  \centering
  \includegraphics[width=\linewidth]{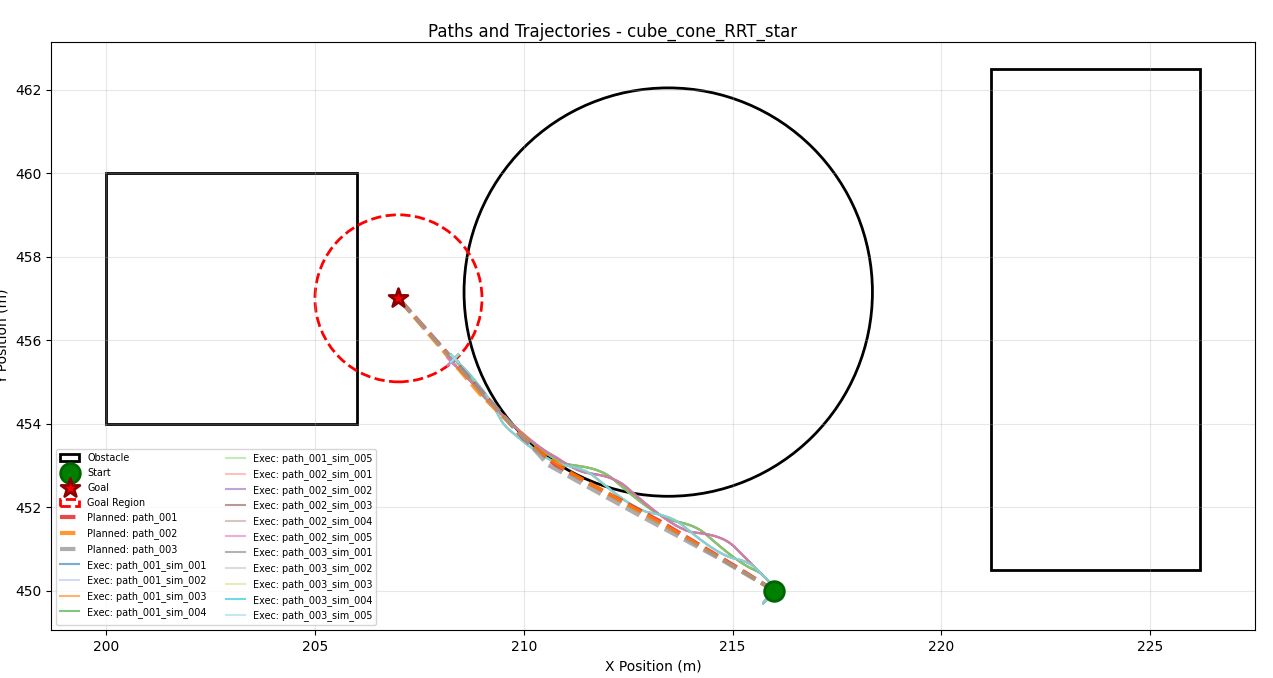}
  \caption{RRT* planner and PD waypoint follower.}
  \label{fig:rrt}
\end{figure}

\begin{table}[h]
\centering
\begin{tabular}{|l|cccc|}
\hline
Scenario & \multicolumn{4}{c}{Trajectory Length (m)} \\
\cline{2-5}
& \textbf{Avg} & \textbf{Std} & \textbf{Min} & \textbf{Max} \\
\hline
Behind Cone Rrt Star & 16.68 & 0.09 & 16.56 & 16.76 \\
Cone Cylinder Rrt Star Level & 8.42 & 0.03 & 8.39 & 8.46 \\
Cube Cone Rrt Star Level & 10.16 & 0.06 & 10.07 & 10.23 \\
Far Behind Cone Rrt Star & 25.29 & 0.17 & 25.13 & 25.53 \\
 
\hline
Behind Cone & 15.67 & 0.00 & 15.67 & 15.67 \\
Cone Cylinder & 8.51 & 0.02 & 8.49 & 8.53 \\
Cube Cone & 10.55 & 0.20 & 10.21 & 10.88 \\
Far Behind Cone & 26.67 & 0.27 & 26.13 & 26.98 \\
\hline
Behind Cone PF Placeholder & 18.80 & 0.03 & 18.77 & 18.87 \\
Cone Cylinder PF & 8.32 & 0.01 & 8.29 & 8.34 \\
Cube Cone PF & 10.65 & 0.01 & 10.64 & 10.67 \\
Far Behind Cone PF placeholder & 26.67 & 0.02 & 25.61 & 26.71 \\
\hline
\hline
\end{tabular}
\caption[Comparison of our method to RRT* + PD controller]{Top half of table:Trajectory Length Statistics for RRT* path and PD controller. Bottom Half: Trajectory Length Statistics for our method.}
\label{tab:trajectory_length}
\end{table}

\subsubsection{Robustness to Noise}
Noise was added to test sensor robustness. Each sensor, except for FLS, is modeled with Gaussian white noise $\eta \sim \mathcal{N}(0, \sigma^2)$ applied to the ground truth measurements. The specific noise standard deviations ($\sigma$) and configurations are detailed in Table \ref{tab:sensor_noise}.

\begin{figure}[htbp]
     \centering
     \begin{subfigure}[b]{0.48\textwidth}
         \centering
         \includegraphics[width=\textwidth]{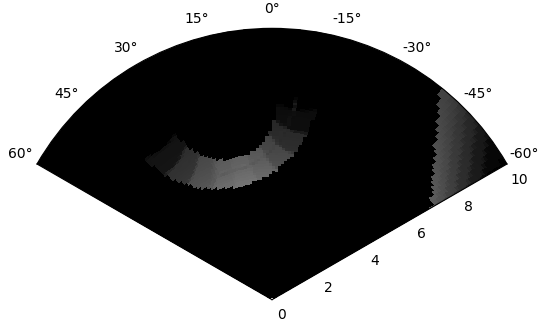}
         \caption{FLS simulation without noise.}
         \label{fig:sonar_no_noise}
     \end{subfigure}
     \hfill
     \begin{subfigure}[b]{0.48\textwidth}
         \centering
         \includegraphics[width=\textwidth]{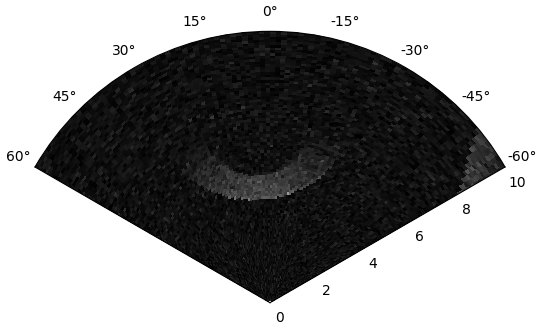}
         \caption{FLS simulation with simulated sensor noise.}
         \label{fig:sonar_noise}
     \end{subfigure}
     \caption[Comparison of Forward-Looking Sonar (FLS)]{Comparison of Forward-Looking Sonar (FLS) output in the simulation environment.}
     \label{fig:sonar_comparison}
\end{figure}

\begin{table}[h]
    \centering
    \caption{AUV Sensor Configuration and Noise Parameters. See \cite{potokar2022holoocean} for information on sonar model}
    \label{tab:sensor_noise}
    \begin{tabular}{@{}llll@{}}
        \toprule
        Sensor & Parameter & Value  \\
        \midrule
        Pose & Position Noise ($\sigma_{pos}$) & 0.1\,m\\
        & Rotation Noise ($\sigma_{rot}$) & 0.0001\,rad \\
        \addlinespace
        IMU & Linear Accel. ($\sigma_{a}$) & 0.012\,\si{\meter/\second^2}  \\
        & Angular Vel. ($\sigma_{\omega}$) & 0.0001\,\si{\radian/\second}  \\
        \addlinespace
        DVL & Velocity Noise ($\sigma_{v}$) & 0.05\,\si{\meter/\second}  \\
        \addlinespace
        Sonar & Additive Sigma ($\sigma_{add}$) & 0.05  \\
        & Multiplicative Sigma ($\sigma_{mult}$) & 0.1 \\
        & Range Sigma ($\sigma_{range}$) & 0.025\,\si{\meter} \\
        & Azimuth Artifacts & 0 \\
        \bottomrule
    \end{tabular}
\end{table}

\paragraph{Inertial and Kinematic Modeling}
The navigation system relies on an Inertial Measurement Unit (IMU) and a Doppler Velocity Log (DVL). The IMU noise is derived from Vectornav V100 sensor, assuming a \SI{10}{\hertz} bandwidth. The acceleration noise is defined as:
\begin{equation}
    \sigma_{accel} = 4\,mg/\sqrt{\text{Hz}} \cdot \sqrt{10\,\text{Hz}} \approx 0.012\,\text{m/s}^2
\end{equation}
while the angular velocity noise is modeled as $\sigma_{\omega} = 1 \times 10^{-4}$\,\text{rad/s}.

The DVL provides body-frame velocity measurements with a standard deviation of $\sigma_{v} = 0.05$\,m/s, simulating acoustic scattering and beam-correlation errors.

\paragraph{Results with sensor noise}
Using the sensor noise values in table \ref{tab:sensor_noise} we found that the agent could deal with added noise without a significant drop in the success rate. For the seen areas, the AUV navigated successfully to the areas to the left and to the right of the cone. For unseen areas, the only marked difference was a slight improvement in the success rate for the area behind the cylinder from 2/20 to 5/20, although this result is not statistically significant.

\begin{table}[htbp]
\centering
\caption{Comparison of Success rates with noise}
\label{tab:tests_noise}
\begin{tabular}{|l|c|c|c|c|}
\hline
\textbf{Area type \textbackslash Area} & \textbf{Near Cube} & \textbf{Behind Cone} & \textbf{Near Cylinder} \\
\hline
Seen goals & 20/20 & 18/20 & 20/20 \\
\hline
Unseen goals & 12/20 & 20/20 & 5/20 \\
\hline
\end{tabular}
\end{table}

\subsubsection{Decreased Visibility}
\begin{figure}[htbp]
     \centering
     \begin{subfigure}[b]{0.45\textwidth}
         \centering
         \includegraphics[width=\textwidth]{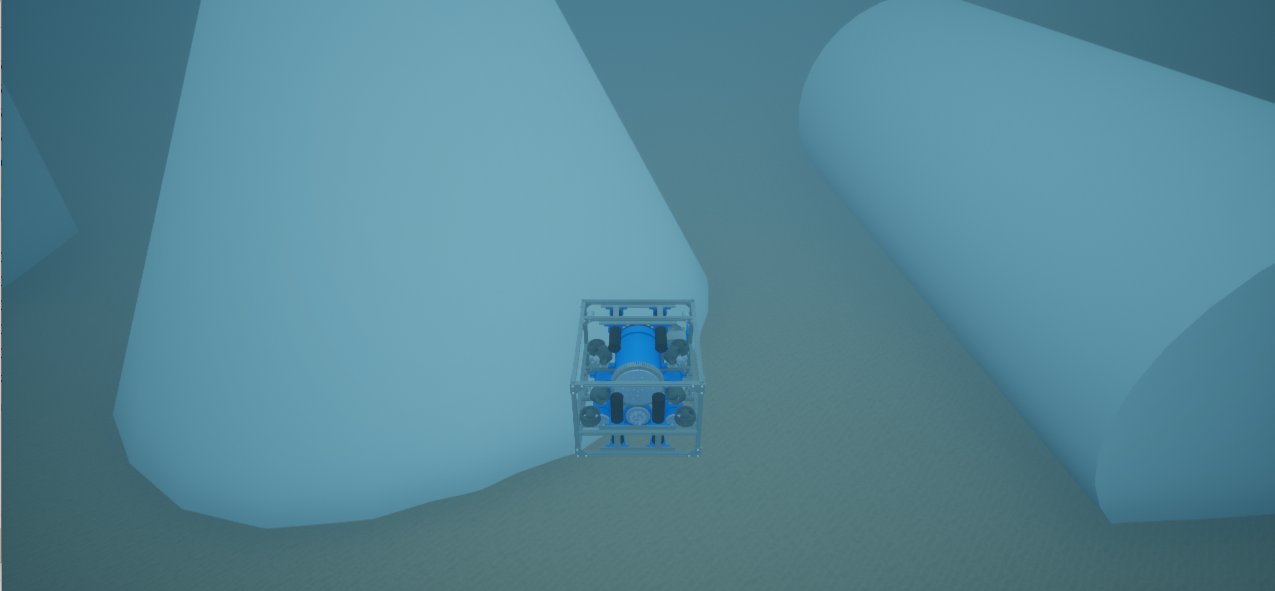}
         \caption{AUV with 0.03 intensity fog}
         \label{fig:AUV_fog_0.03}
     \end{subfigure}
     \hfill 
     \begin{subfigure}[b]{0.45\textwidth}
         \centering
         \includegraphics[width=\textwidth]{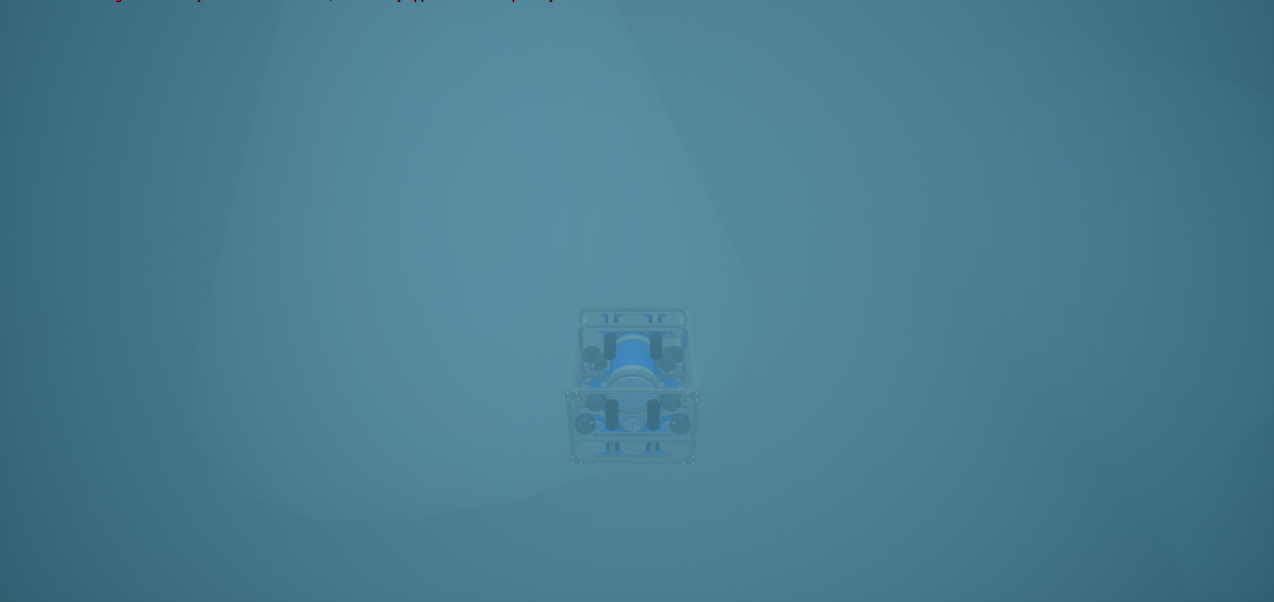}
         \caption{AUV with 0.1 intensity fog}
         \label{fig:right_side}
     \end{subfigure}
     
     \caption{Comparison of two levels of fog}
     \label{fig:fog_comparison}
\end{figure}

In another challenge to our agent, we decreased the visibility and added noise to all sensors except the FLS. This was done by increasing fog level in the simulator from 0.03, seen in training, to 0.1. The difference in fog can be viewed in figure \ref{fig:fog_comparison}. We stress that the agent has never encountered seen fog at these levels, nevertheless, there is no drop in performance in the seen areas. In unseen areas, the results were mixed, in one area the success rate increased from 2/20 without sensor noise to 8/20 in the current configuration. In another area the success rate decreased from 12/20 to 8/20.

\begin{table}[htbp]
\centering
\caption{Comparison of Success rates with sensor noise (excluding sonar noise) and fog}
\label{tab:tests_fog}
\begin{tabular}{|l|c|c|c|c|}
\hline
\textbf{Area type \textbackslash Area} & \textbf{Near Cube} & \textbf{Behind Cone} & \textbf{Near Cylinder} \\
\hline
Seen goals & 22/22 & 22/22 & 22/22 \\
\hline
Unseen goals & 8/20 & 22/22 & 8/20 \\
\hline
\end{tabular}
\end{table}

\begin{figure}[h]
  \centering
  \includegraphics[width=\linewidth]{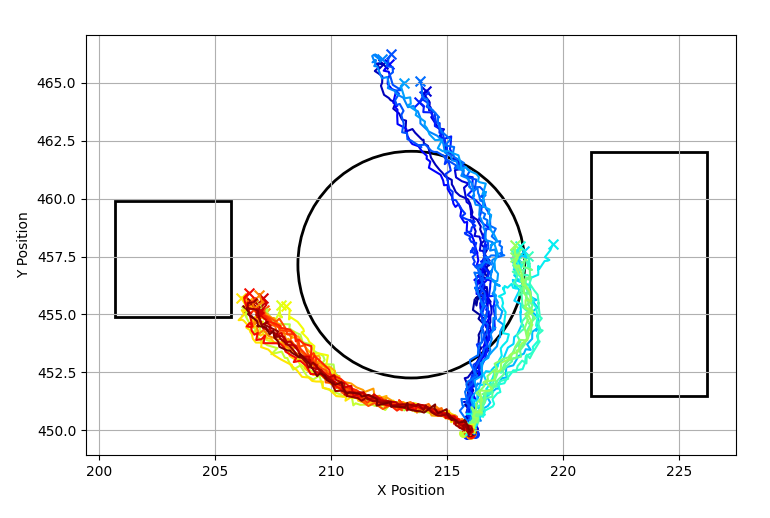}
  \caption{Test with goals in seen areas and sensor noise.}
  \label{fig:seen_noise}
\end{figure}

\section{Discussion}
The results of this study demonstrate that an end-to-end deep reinforcement learning (DRL) approach is capable of achieving underwater obstacle avoidance using raw sensor data. While the current High Level Reinforcement Learning implementation did not outperform the traditional sampling-based $RRT^{*} + PD$ pipeline, the performance gap was remarkably narrow. Specifically, the RL policy achieved trajectory lengths within 4\% of the RRT* baseline in areas seen during training and within 6\% in novel areas. This suggests that the DRL policy learned an efficient navigation strategy that closely approximates the mathematical optimality of sampling-based planners. When adding noise to the sensor input, the success rates of the policy did not change, showing that the network architecture is robust to sensor noise.
\subsection{Advantages of the Proposed Method}
The primary advantage of this approach is the reduction of manual engineering complexity. Traditional AUV pipelines require separate, highly configured modules for mapping, path planning, and control. In contrast, the proposed DRL framework maps raw sonar, visual, and proprioceptive data directly to thruster commands, significantly simplifying the deployment process. Furthermore, the policy showed promising generalization capabilities, successfully navigating unvisited areas by identifying and avoiding obstacles similar to those encountered during training
\subsection{Robustness and Generalization}
Our experiments validate that the robustness to visual perturbations reported in the original SERL framework  extends effectively to the underwater domain. Despite being trained on idealized, noise-free sensor data, the policy maintained high success rates when subjected to distinct environmental challenges. Specifically, performance remained stable in seen areas both when Gaussian noise was introduced to the sensors and, in separate tests, when visibility was significantly reduced by fog.

The failure to generalize to the cylinder geometry highlights a 'brittleness' in the learned perception-to-action mapping when encountering novel geometric primitives. The agent’s tendency to enter a local minimum  indicates that the policy's value function likely overestimated the safety of the proximity to the cylinder's curved surface. Interestingly, the marginal improvement in success rates under noisy conditions (from 5  to 25 and 40 percent with fog)  supports the hypothesis that the baseline policy was trapped in a suboptimal attractor state. The stochastic perturbations introduced by sensor noise provided sufficient exploration to occasionally bypass the obstacle, yet the fundamental lack of geometric invariance in the visual encoder remains the primary bottleneck for universal obstacle avoidance.

However, generalization to novel geometries remains a challenge. While the policy successfully navigated unvisited areas that shared geometric similarities with the training set (e.g., "far behind cone") , it failed in the "near cylinder" scenario, achieving only a 2/20 success rate in baseline tests. Interestingly, the addition of sensor noise and fog marginally increased this success rate to 5/20 and 8/20, respectively. We interpret this not as a functional improvement, but as confirmation that the agent was trapped in a local minimum ; the added noise provided enough stochastic perturbation to occasionally dislodge the agent, but the consistently low success rate confirms that the policy failed to fundamentally generalize to the cylinder's end-cap geometry.

\subsection{Limitations and Challenges}
Despite these advantages, the study highlighted several limitations. First, the performance of the policy degraded in unvisited environments where the obstacle geometry differed significantly from the training set—specifically for the cylinder obstacle where success rates dropped to $2/20$. This indicates that the policy's ability to generalize is currently tied to the diversity of the training data. Second, while the SERL framework is designed for sample efficiency, our setting required more demonstrations and longer training times than original benchmarks. We attribute this to the high complexity of the underwater visual environment, where lighting and texture change dramatically compared to static tabletop robotic environments.

\section{Conclusion and Future Work}
This research presented an end-to-end hierarchical reinforcement learning framework for the autonomous navigation and obstacle avoidance of underwater vehicles. By decoupling motion execution from motion planning through a dual MDP structure, we successfully trained a policy that ingests raw $84 \times 84$ pixel camera frames and $100 \times 100$ pixel sonar images to output direct thruster commands.

Our experiments confirm that the trained policy achieves high success rates in navigating around geometric obstacles, with efficiency comparable to a standard RRT* planner. While the system demonstrated robustness to simulated sensor noise and decreased visibility (fog), it struggled to generalize to obstacle geometries significantly different from its training set, specifically cylindrical shapes. Furthermore, although the SERL framework enabled learning from demonstrations, the visual and environmental variability of the underwater domain necessitated a higher volume of expert data compared to terrestrial robotics tasks.

Our framework can also be adapted to other tasks. We assume that it would be easier to adapt to a narrower task, such as docking where the station stays relatively the same and has visual aids. Another task worth exploring would be interception and following of different vehicle types in open water. Other future work would include testing out our proposed method on different vehicle types, in different water types and, of course, on a real vehicle at sea.

\bibliographystyle{IEEEtran}
\bibliography{references,trad_nav_pipe,end_to_end, uw_perception}

\end{document}